\documentclass[10pt,twocolumn,letterpaper]{article}

\usepackage[pagenumbers]{cvpr} 

\usepackage{graphicx}
\usepackage{amsmath}
\usepackage{amssymb}
\usepackage{booktabs}

\usepackage{siunitx}
\usepackage{cite}
\usepackage{ctable}
\usepackage{hhline}
\usepackage{booktabs}
\usepackage{subcaption}
\usepackage{csquotes}
\usepackage{multirow}

\usepackage{cuted}
\usepackage{capt-of}

\usepackage{color, colortbl}
\definecolor{Gray}{gray}{0.9}

\newcolumntype{L}[1]{>{\raggedright\let\newline\\\arraybackslash\hspace{0pt}}m{#1}}
\newcolumntype{C}[1]{>{\centering\let\newline\\\arraybackslash\hspace{0pt}}m{#1}}
\newcolumntype{R}[1]{>{\raggedleft\let\newline\\\arraybackslash\hspace{0pt}}m{#1}}

\usepackage{comment}

\usepackage[pagebackref,breaklinks,colorlinks]{hyperref}

\usepackage[capitalize]{cleveref}
\crefname{section}{Sec.}{Secs.}
\Crefname{section}{Section}{Sections}
\Crefname{table}{Table}{Tables}
\crefname{table}{Tab.}{Tabs.}

\def\cvprPaperID{9866} 
\def\confName{CVPR}
\def\confYear{2022}

\begin{document}

\title{Camera Motion Agnostic 3D Human Pose Estimation}

\iftrue
\author{
Seong Hyun Kim$^1$ \hspace{1.0cm} Sunwon Jeong$^2$ \hspace{1.0cm} Sungbum Park$^2$ \hspace{1.0cm} Ju Yong Chang$^1$\\
$^1$Dept of ECE, Kwangwoon University, Korea\hspace{1.0cm}
$^2$NCSOFT, Korea\\
{\small \texttt {\{thuthdew15,jychang\}@kw.ac.kr},\ \ \texttt {\{bmycrew,spark0916\}@ncsoft.com}}
}
\fi

\maketitle

\begin{strip}
\centering
\includegraphics[width=\linewidth]{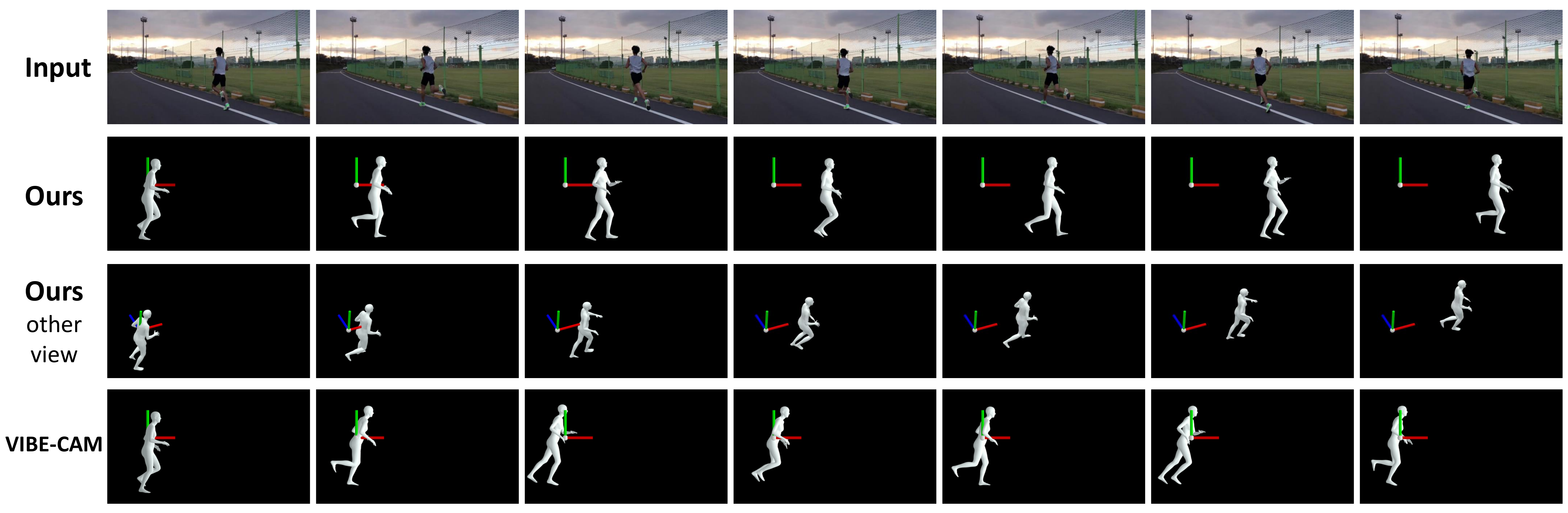}
\vspace*{-6mm}
\captionof{figure}{Given a runner video (first row), the proposed framework correctly reconstructs 3D running path (second and third rows), while VIBE-CAM, the combination of state-of-the-art human pose estimation methods~\cite{kocabas2020vibe, habermann2020deepcap, raaj2019efficient}, fails to reconstruct the 3D global pose of the runner (fourth row). The global pose represents the orientation and location of the entire body. The visualized reference frame is defined as being aligned with the person in the first frame. VIBE-CAM is detailed in Sec.~\ref{subsection: VIBE-CAM detail}.
\label{fig1}}
\vspace*{-1mm}
\end{strip}

\begin{abstract}
Although the performance of 3D human pose and shape estimation methods has improved significantly in recent years, existing approaches typically generate 3D poses defined in camera or human-centered coordinate system. This makes it difficult to estimate a person's pure pose and motion in world coordinate system for a video captured using a moving camera. To address this issue, this paper presents a camera motion agnostic approach for predicting 3D human pose and mesh defined in the world coordinate system. The core idea of the proposed approach is to estimate the difference between two adjacent global poses (i.e., global motion) that is invariant to selecting the coordinate system, instead of the global pose coupled to the camera motion. To this end, we propose a network based on bidirectional gated recurrent units (GRUs) that predicts the global motion sequence from the local pose sequence consisting of relative rotations of joints called global motion regressor (GMR). We use 3DPW and synthetic datasets, which are constructed in a moving-camera environment, for evaluation. We conduct extensive experiments and prove the effectiveness of the proposed method empirically. Code and datasets are available at \url{https://github.com/seonghyunkim1212/GMR} 




\end{abstract}

\section{Introduction}

3D human pose estimation~\cite{huang2017towards,pavlakos2017coarse,martinez2017simple,pavllo20193d,kanazawa2018end,kolotouros2019learning,kocabas2019self,guler2019holopose,kolotouros2019convolutional,Moon_2020_ECCV_I2L-MeshNet,kocabas2020vibe} is an important topic in computer vision that can be applied to many applications, such as virtual/augmented reality, human action recognition, and human behavior understanding. The performance of 3D human pose estimation has improved remarkably thanks to advances in deep learning.

\begin{figure*}[t]
\includegraphics[width=\textwidth]{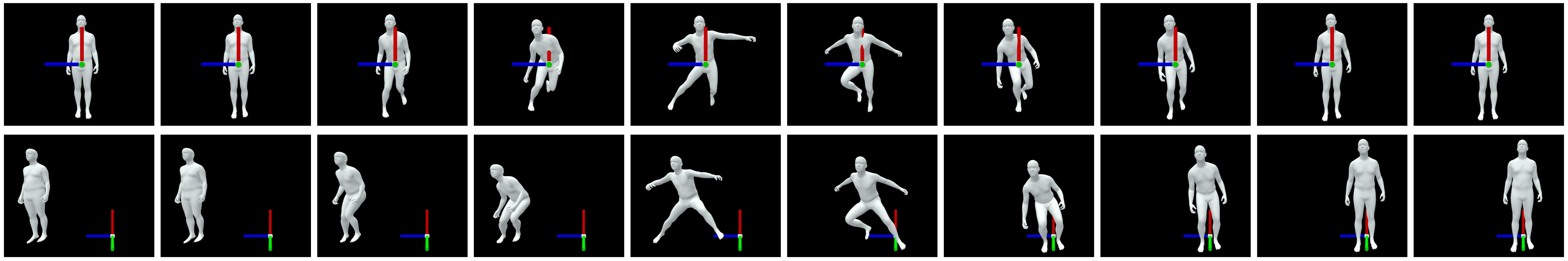}
\vspace*{-6mm}
\caption{The top row shows the image sequence rendered using only the local pose without the global pose. Here, the relative orientations between rigid body parts (i.e., local pose) change, but the entire body's orientation and location (i.e., global pose) remain unchanged. The bottom row shows the rendering result for the case where the global pose is further included. Please note that the main purpose of the paper is to estimate the global pose sequence from the local pose sequence.}
\label{fig2}
\vspace*{-1mm}
\end{figure*}

Most 3D human pose estimation methods reconstruct 3D poses defined in the camera or human-centered coordinate system. The estimated 3D human pose is coupled to the camera pose. Therefore, reconstructing intrinsic human poses for a video sequence captured by a moving camera is challenging. Our paper addresses this problem and proposes a method to estimate the \emph{intrinsic human pose} independent of camera motion. Fig.~\ref{fig1} shows the difference between 3D human pose sequences reconstructed using the proposed and existing methods.

In the kinematic chain model for human body or the statistical human shape model, such as SMPL~\cite{loper2015smpl}, 3D human poses can be decomposed into a \emph{local pose} that represents the orientation of rigid body parts and a \emph{global pose} that represents the orientation and location of the entire body, as shown in Fig.~\ref{fig2}. The local pose is represented hierarchically through relative rotations of rigid body parts from the rest pose (i.e., zero pose) and defined in the generic coordinate system~\cite{loper2015smpl}. Therefore, the local pose is independent of the selection of the reference coordinate system. However, the global pose is dependent on the selection of the reference coordinate system. The global pose is generally defined on the basis of the camera coordinate system in existing methods~\cite{pavlakos2018learning,kanazawa2018end,kolotouros2019learning,kocabas2020vibe,Luo_2020_ACCV}; thus, the estimated 3D human pose is coupled to the camera motion. Our basic idea is to estimate the difference in the global pose in adjacent frames (i.e., global motion) invariant to the selection of the reference coordinate system instead of the global pose coupled to the camera motion.

So, how can we estimate the global motion decoupled from the camera motion? Our intuition is that the \emph{global motion} (i.e., global pose displacement between neighboring frames) can be predicted from the local pose sequence, as shown in Fig.~\ref{fig2}. Suppose a person makes a jump to the left. We can easily infer that the person jumps to the left as shown in Fig.~\ref{fig2} (bottom row) from the local pose sequence in Fig.~\ref{fig2} (top row). Therefore, our goal is to design a deep network that estimates the global human motion sequence from the local human pose sequence. Specifically, the local pose sequence is reconstructed from a video using an existing 3D human pose estimation method, such as VIBE~\cite{kocabas2020vibe}. We model the mapping function from the input local pose sequence to the output global motion sequence through a temporal network called global motion regressor (GMR) and train the network using the large-scale motion capture dataset AMASS~\cite{mahmood2019amass}.

We evaluate the proposed method using the 3DPW dataset~\cite{von2018recovering}. We also synthesize an animated 3D human pose dataset using CMU sequences in the AMASS dataset to allow camera movement in the synthetic video. Both datasets are used for qualitative and quantitative evaluations of the proposed method.



The main contributions of this paper are presented as follows:
\begin{itemize}
\item To the best of our knowledge, our work is the first deep learning-based framework for predicting a pure human pose independent of camera motion. We demonstrate that it is possible to estimate the human pose sequence in the world coordinate system without camera calibration from a video including camera motion.
\item We propose a model based on gated recurrent units (GRUs)~\cite{cho-etal-2014-learning} that transforms the local human pose sequence into the global motion sequence invariant to the selection of the reference coordinate system. The proposed model can be combined with any human pose estimation method that predicts local human poses.
\item We propose new metrics for the evaluation of the proposed method. Moreover, we train the proposed model for various input/output rotation representations and rotation loss functions and quantitatively compare them using the proposed evaluation metrics to determine the optimal rotation representation and loss function.
\end{itemize}

\section{Related Works}



Methods for simultaneously reconstructing 3D human poses and shapes are reviewed in this section.

\begin{figure*}[t]
\includegraphics[width=\linewidth]{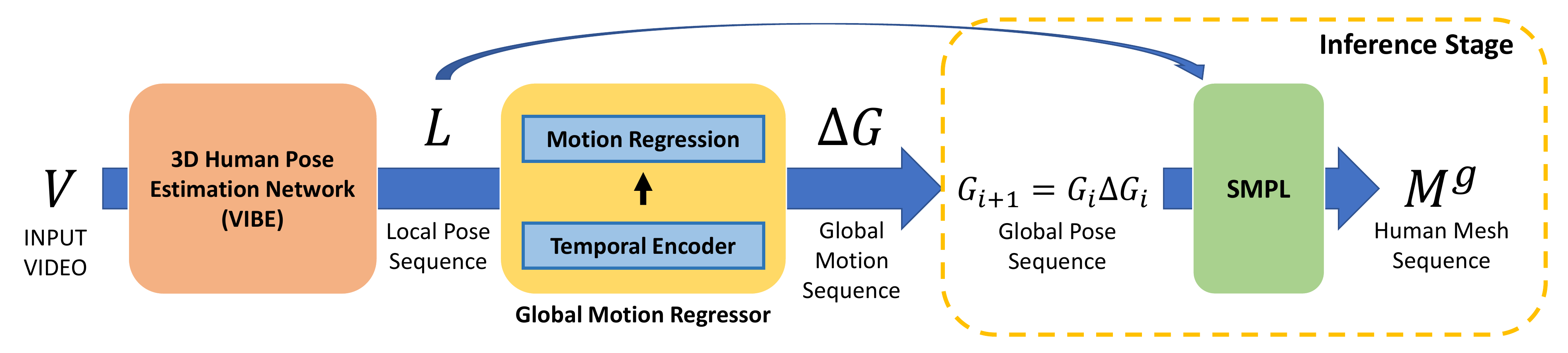}
\vspace*{-6mm}
\caption{\textbf{Overall framework of the proposed method.} Given an input video, the existing 3D human pose estimation network outputs a local human pose sequence. The proposed global motion regressor generates a global motion sequence from the local pose sequence. In the inference stage, the global motion is accumulated into a global pose, and finally, the SMPL reconstructs a human mesh sequence with the global pose defined in the world coordinate system.}
\label{fig3}
\vspace*{-1mm}
\end{figure*}

\textbf{3D human pose and shape estimation from a single image.} The method for estimating the 3D human pose and shape from a single image can be divided into model-based and model-free approaches. Model-based approaches commonly use the statistical body shape model SMPL~\cite{loper2015smpl} to reconstruct the human shape and allow the network to predict parameters of the SMPL model. Meanwhile, the model-free approach performs 3D human shape reconstruction by directly estimating a 3D human mesh instead of predicting SMPL parameters. \cite{kanazawa2018end,pavlakos2018learning,omran2018neural,kolotouros2019learning} belong to the model-based approach. Kanazawa~\textit{et al.}~\cite{kanazawa2018end} introduced an adversarial training method to obtain an anthropometrically plausible 3D shape and proposed a discriminator network. Pavlakos~\textit{et al.}~\cite{pavlakos2018learning} used keypoints and silhouettes as an intermediate representation for predicting SMPL parameters. Omran~\textit{et al.}~\cite{omran2018neural} utilized body part segmentation. Kolotouros~\textit{et al.}~\cite{kolotouros2019learning} proposed a method that combines feedforward regression step and SMPLify-based optimization step~\cite{bogo2016keep} into a loop structure to combine advantages of regression-based and optimization-based methods. \cite{varol2018bodynet,kolotouros2019convolutional,Moon_2020_ECCV_I2L-MeshNet,lin2021end} belong to the model-free approach. Varol~\textit{et al.}~\cite{varol2018bodynet} proposed a network that directly predicts a 3D human mesh in volumetric space and used keypoints, segmentation, and 3D pose as the intermediate representation for this. Kolotouros~\textit{et al.}~\cite{kolotouros2019convolutional} proposed a graph convolutional network for 3D human mesh reconstruction. Their network takes rest poses and image features as inputs and directly regresses the 3D human mesh. Moon~\textit{et al.}~\cite{Moon_2020_ECCV_I2L-MeshNet} proposed the image-to-lixel prediction network that predicts vertex coordinates of the 3D human mesh through 1D heatmaps. Lin~\textit{et al.}~\cite{lin2021end} recently proposed a transformer-based network that simultaneously reconstructs human pose and shape by modeling vertex-vertex and vertex-joint interactions.

\textbf{3D human pose and shape estimation from a video.} Kanazawa~\textit{et al.}~\cite{kanazawa2019learning} proposed a method for predicting not only 3D meshes that correspond to a single input image but also those that correspond to frames in the past and future through learning using video data. Arnab~\textit{et al.}~\cite{arnab2019exploiting} proposed a bundle-adjustment-based algorithm that temporally and consistently refines initial per-frame SMPL estimates. Sun~\textit{et al.}~\cite{sun2019dsd-satn} proposed a transformer-based temporal model. In that study, in order for the network to learn temporal information better, the order of shuffled frames can be predicted, and an unsupervised adversarial training method for this was proposed. Kocabas~\textit{et al.}~\cite{kocabas2020vibe} proposed a temporal model based on GRU. In that study, a motion discriminator network was proposed to allow the network to generate a plausible 3D human motion. Luo~\textit{et al.}~\cite{Luo_2020_ACCV} proposed a two-stage model for human motion estimation. Overall coarse motion is predicted using variational motion estimation in the first stage of the model and then further improved through motion residual regressor in the second stage. Choi~\textit{et al.}~\cite{choi2021beyond} proposed a method that reconstructs a temporally consistent human shape using temporal information of past and future frames.

All the methods reviewed above reconstruct 3D human pose and shape in the camera coordinate system. The result reconstructed by these methods from a video captured by a fixed camera can be considered to be defined in the world coordinate system. However, assuming the general environment where no extrinsic camera parameters are given, it is difficult to convert the reconstruction result from a video with camera motion into the pure 3D human pose defined in the world coordinate system. One possible method is to extract camera motions from the video using a structure-from-motion (SfM) method such as COLMAP~\cite{schonberger2016structure}, and use them to transform the human pose in the camera coordinate system into the world coordinate system. However, SfM methods often fail to achieve successful results in videos containing dynamic objects. Although foreground-background segmentation can be used for removal of dynamic foreground objects, a fully automated method for camera motion estimation is still unavailable. Our goal is to obtain the pure 3D human pose sequence in the world coordinate system without camera calibration from a video with any camera motion. As far as we know, there is no deep-learning-based method to reconstruct 3D human poses in the world coordinate system from a video captured by a moving camera, and we propose this method for the first time in this paper.

\section{Proposed Method}

\subsection{Overall Approach}

Fig.~\ref{fig3} shows the overall framework of the proposed method. First, we use a human pose estimation network to determine the local pose sequence $L=\{L_{i}\}^{T}_{i=1}$ given an input video $V=\{V_{i}\}^{T}_{i=1}$ with length $T$, where $L_{i}\in{\mathbb{R}^{92}}$ represents the relative rotations of 23 joints in an unit-quaternion form. Second, bidirectional GRU-based temporal encoder outputs a latent feature containing temporal information of this sequence from the local pose sequence $L$. We obtain the global motion sequence $\Delta G=\{\Delta G_{i}\}^{T}_{i=1}$ from the latent feature through the motion regression layer. A global motion $\Delta G_{i}$ consists of an orientation motion $\Delta A_{i}\in{\mathbb{R}^{3}}$ in an axis-angle form and a translation motion $\Delta T_{i}\in{\mathbb{R}^{3}}$. Third, we accumulate estimated global motions starting with an initial global pose to compute a global pose sequence $G=\{G_{i}\}^{T}_{i=1}$. Finally, the computed global pose sequence $G$ and the input local pose sequence $L$ are converted into a global human mesh sequence $M^{g}=\{M^{g}_{i}\}^{T}_{i=1}$ defined in the world coordinate system through the SMPL model~\cite{loper2015smpl}.

\subsection{SMPL Representation}

SMPL~\cite{loper2015smpl} represents human pose and shape using the pose parameter $\theta\in{\mathbb{R}^{72}}$ and the shape parameter $\beta\in{\mathbb{R}^{10}}$. The pose parameter is parameterized by global 3D rotation and the relative 3D rotations of 23 joints in an axis-angle representation. The shape parameter is parameterized using the first 10 principal component coefficients of the human shape space. SMPL provides a differentiable function that generates the 3D human mesh $M(\theta,\beta)\in{\mathbb{R}^{6890\times3}}$ from the pose parameter $\theta$ and the shape parameter $\beta$. Relative rotations of the 23 joints of the pose parameter that correspond to the local pose become the input to GMR. However, since GMR uses the local pose represented in an unit quaternion form as an input, the local pose represented in the unit quaternion form is first transformed to an axis-angle form, which is then used as an input of the SMPL model. Global rotation corresponds to the global pose's orientation, which is the output of GMR. In this work, the shape parameter is obtained using the existing 3D human pose estimation method~\cite{kocabas2020vibe}. Unlike existing methods~\cite{pavlakos2018learning,kanazawa2018end,kolotouros2019learning,kocabas2020vibe,Luo_2020_ACCV}, the proposed method generates a global human mesh defined in the world coordinate system by adding the translation to the 3D human mesh $M$ as follows:
\begin{equation}
\label{eq:human mesh with translation}
    {M^{g}}(\theta,\beta,T)=M(\theta,\beta)+T,
\end{equation}
where $T\in{\mathbb{R}^{3}}$ denotes the global translation, which is one of the outputs of the proposed method.

\begin{figure}[t]
\includegraphics[width=\linewidth]{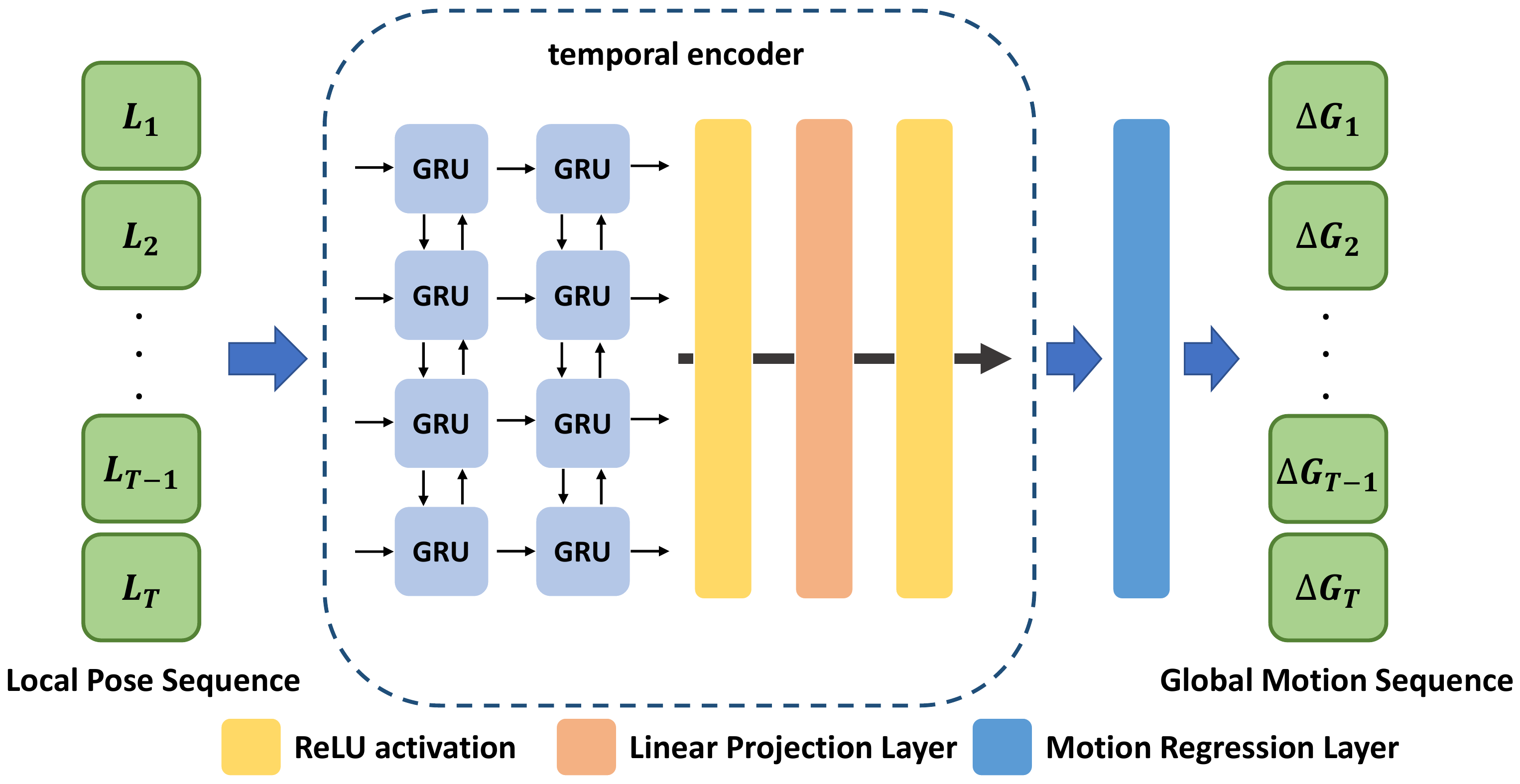}
\vspace*{-6mm}
\caption{Architecture of \textbf{G}lobal \textbf{M}otion \textbf{R}egressor (GMR).}
\label{fig4}
\vspace*{-1mm}
\end{figure}

\subsection{Global Motion Regressor (GMR)}


The proposed network estimates the global motion sequence, that is, the deviation of global poses between two adjacent frames from the local pose sequence $L$. Various temporal neural architectures have been proposed to address these types of sequence data in recent years. We model GMR using bidirectional GRU~\cite{cho-etal-2014-learning} to encode long-term information effectively in this work. Fig.~\ref{fig4} shows the architecture of the proposed GMR network. First, the local pose sequence $L=\{L_{i}\}^{T}_{i=1}$ is fed into the temporal encoder that consists of bidirectional GRUs and a linear projection layer. Each bidirectional GRU forwards the local pose sequence to the GRU layer in forward and reverse directions and concatenates their results to generate hidden states $H=\{H_{i}\}^{T}_{i=1}$, where $H_{i}\in{\mathbb{R}^{4096}}$. Second, the dimension of output hidden states $H$ is reduced by the linear projection layer and the linear projection layer then generates the latent feature $F=\{F_{i}\}^{T}_{i=1}$, where $F_{i}\in{\mathbb{R}^{2048}}$. Finally, the motion regression layer regresses the global motion sequence $\Delta{G}=\{\Delta{G}_{i}\}^{T}_{i=1}$ from the latent feature $F$. The global motion $\Delta{G_{i}}=(\Delta{A_{i}},\Delta{T_{i}})$ consists of orientation $\Delta{A_{i}}$ and translation $\Delta{T_{i}}$ motions between $i$-th and $(i+1)$-th frames. $\Delta{A_{i}}$ represented in the axis-angle form is transformed to a 3$\times$3 rotation matrix $\Delta{R_{i}}$ through the Rodrigues' rotation formula~\cite{gallego2015compact}. Then, the global motion $\Delta{G_{i}}\in{SE(3)}$ can be written using $\Delta{R_{i}}\in{SO(3)}$ and $\Delta{T_{i}}\in{\mathbb{R}^3}$. Moreover, the global pose $G_{i}\in{SE(3)}$ can be represented using $R_{i}\in{SO(3)}$ and $T_{i}\in{\mathbb{R}^3}$. The following equations hold between the global pose $G_{i}$ and the global motion $\Delta{G_{i}}$:
\begin{equation}
\label{eq: global pose calculation}
    {G_{i+1}}={G_{i}}{\Delta{G_{i}}}=
    \begin{bmatrix}
    R_{i} & T_{i} \\
    \mathbf{0}^{T} & 1 \\
    \end{bmatrix}
    \begin{bmatrix}
    \Delta{R_{i}} & \Delta{T_{i}} \\
    \mathbf{0}^{T} & 1 \\
    \end{bmatrix}
    ,
\end{equation}
\begin{equation}
\label{eq: orientation calculation}
    R_{i+1}=R_{i}\Delta{R_{i}},
\end{equation}
\begin{equation}
\label{eq: translation calculation}
    T_{i+1}=R_{i}\Delta{T_{i}}+T_{i}.
\end{equation}
Finally, through the SMPL model~\cite{loper2015smpl}, we reconstruct the global human mesh ${M}^{g}_{i}={M}^{g}([A_{i},L_{i}],\beta_{i},T_{i})$ defined in the world coordinate system from the obtained global poses $G_{i}$ and input local poses $L_{i}$, where $A_{i}$ is the axis-angle form of $R_{i}$ and $[\cdot,\cdot]$ denotes the concatenation.

\subsection{Loss Function}
\label{subsection: loss function}

We train the proposed GMR using the following loss function:
\begin{equation}
\begin{split}
\label{eq: total loss function}
    L_{total}=w_{ori}L_{ori}+w_{trans}L_{trans}\\
    +w_{vertex}L_{vertex}+w_{smooth}L_{smooth},
\end{split}
\end{equation}
where $L_{ori}$, $L_{trans}$, $L_{vertex}$, and $L_{smooth}$ are orientation, translation, vertex, and smoothness losses, respectively; and $w_{ori}$, $w_{trans}$, $w_{vertex}$, and $w_{smooth}$ denote weights of losses and are set to $1$, $1$, $1$, and $10^{-2}$, respectively. We must carefully define the loss function $L_{ori}$ to supervise the predicted orientation motion $\Delta{R_{i}}$ because the 3D rotation belongs to $SO(3)$, not the Euclidean space. Accordingly, we test angular $L_{angular}$, chordal $L_{chordal}$, and axis-angle $L_{axis\text{-}angle}$ losses, which are based on commonly used distance measures for $SO(3)$~\cite{hartley2013rotation}. We also define the translation loss $L_{trans}$ using the Euclidean distance between the predicted translation motion $\Delta{T_{i}}$ and its ground-truth. For vertex-wise loss on the reconstructed 3D mesh surface, we further define the vertex loss $L_{vertex}$ based on the L1 distance. GMR predicts the global motion, which is the temporal deviation of global poses between two adjacent frames. Therefore, instead of directly supervising the global human mesh $M^{g}_{i}$, we apply the vertex loss to the global human mesh offset $\Delta{M}^{g}_{i}=M^{g}([\Delta{A}_{i},L_{i}],\beta_{i},\Delta{T}_{i})$. Finally, we use the smoothness loss $L_{smooth}$ to generate a smooth global motion. It is based on the Frobenius norm between orientation motions in adjacent frames and helps to reconstruct temporally coherent global orientations. Details of all loss functions are provided in the supplementary material.

\subsection{Flip Augmentation}

We use the large-scale motion capture dataset AMASS~\cite{mahmood2019amass} to train the proposed GMR. The AMASS dataset provides large amounts of sequence data from a wide range of human actions. However, its diversity is still limited compared with the variation of real human action. Therefore, we randomly flip sequences of the AMASS dataset in the temporally reverse direction and use them for learning. The used data augmentation process allows the network to utilize additional diverse training data. In this work, we call it flip augmentation, which uses both original and flipped datasets for training.

\subsection{Inference}
\label{subsection: inference}

Given an input video of the frame length $T$, we estimate the local pose sequence $L$ using the existing human pose estimation network~\cite{kocabas2020vibe}. GMR then estimates the global motion sequence $\Delta{G}$ from $L$. We assume that a person moves from the origin of the world coordinate system, and the orientation and the translation of the initial global pose $G_{1}$ are defined as an identity matrix and a 3D zero vector, respectively. Thus, the initial global human pose is represented as $G_{1}=I_{4\times{4}}$.
The global human pose sequence $G=\{G_{1},\ldots,G_{T}\}$ is subsequently calculated by repeatedly applying Eqs.~\ref{eq: orientation calculation} and~\ref{eq: translation calculation} starting with the initial global human pose $G_{1}$. Finally, we put the global human pose sequence $G$ and the local pose sequence $L$ into SMPL~\cite{loper2015smpl} and obtain the global human mesh sequence $M^g=\{M^g_1,\ldots,M^g_T\}$ defined in the world coordinate system using Eq.~\ref{eq:human mesh with translation}.

\section{Experimental Results}

\subsection{Implementation Details}

We set the sequence length and frame rate of the input video to 64 and 10 fps, respectively, to train GMR. However, GMR can work for input sequences of arbitrary length. We use VIBE~\cite{kocabas2020vibe} in the test stage to obtain the local pose sequence. VIBE outputs SMPL pose parameters consisting of global orientations and local poses. However, we only use the local pose from the VIBE output, discard the global orientation, and reconstruct the new global orientation using the proposed method. This is because the global orientation generated by VIBE is defined in the camera coordinate system, so it fails to provide a 3D human pose in the world coordinate system. The bidirectional GRU of the temporal encoder consists of four layers with 2048 neurons, and the linear projection layer consists of one linear layer with 2048 neurons. The motion regression layer consists of one linear layer that outputs the global motion. We use the Adam optimizer~\cite{kingma2014adam} to optimize the loss function and set the learning rate to $5\times{10}^{-5}$. We set the mini-batch size to 32 and train the network using one Nvidia RTX3090 GPU.

\subsection{Datasets}
\label{subsection: datasets}

We use the AMASS~\cite{mahmood2019amass} dataset for training. The AMASS dataset consists of sequences of publicly available datasets, such as CMU MoCap~\cite{de2009guide} and TotalCapture~\cite{trumble2017total}, and provides SMPL parameters extracted using MoSh++. We sample each sequence of the AMASS dataset at a rate of 10 fps and use them for training.

We use three datasets for evaluation. The first dataset, Human3.6M~\cite{ionescu2013human3}, is widely used in 3D human pose estimation research. The Human3.6M dataset provides 3.6M video frames composed of images captured from fixed cameras. We use SMPL parameters extracted via MoSh~\cite{loper2014mosh} for quantitative evaluation, and S9 and S11 of seven subjects are used for evaluation. We use the Human3.6M for ablation experiments and utilize the ground-truth local pose sequence as the input to GMR in this case.

The second dataset, 3DPW~\cite{von2018recovering}, contains 60 sequences captured outdoors. The 3DPW dataset provides global human pose information, but it is difficult to use for evaluation due to severe drift. Therefore, we use the existing SfM method, COLMAP~\cite{schonberger2016structure}, to obtain the camera pose of the 3DPW dataset and generate the pseudo ground-truth global human pose from the obtained camera pose. Simply applying COLMAP fails to obtain successful results because the 3DPW dataset contains dynamic objects. We mask out dynamic objects such as people, vehicles, and animals using Mask R-CNN~\cite{he2017mask} so that COLMAP extracts features only from static regions. Also, we manually filter sequences with poor estimation results. As a result, we obtain global human poses for 17 sequences and perform evaluation on these sequences. Please refer to the supplementary material for the detailed process.

Although the 3DPW dataset consists of various scenes, the camera motion is limited. Therefore, we additionally build animated synthetic videos with circular camera motions using SMPL parameters of the AMASS dataset and use them for evaluation. It is more challenging than linear or panning motions. We use CMU sequences of the AMASS to create these synthetic videos. The CMU dataset of AMASS consists of 106 subjects. We use 50 sequences for 16 subjects to create synthetic videos, and the remaining sequences are included in the training set. The detailed process of constructing our synthetic dataset is provided in the supplementary material.




\subsection{Evaluation Metrics}


Our proposed method predicts global human motion to obtain the intrinsic 3D human pose decoupled from camera motion. As far as we know, there is no metric for quantitatively evaluating the estimated global motion by the proposed method. We propose new metrics to evaluate the proposed method. The first metric is the orientation motion error (OME), which computes the angular error between the estimated global orientation motion $\Delta{A}$ and its ground-truth $\Delta{A}^{*}$. Next, we employ the translation motion error (TME), which computes the Euclidean distance between the predicted translation motion $\Delta{T}$ and its ground-truth $\Delta{T}^{*}$. Finally, the vertex motion error (VME) calculates the vertex-wise Euclidean distance between the prediction $\Delta{M}^{g}$ and ground-truth ${\Delta{M}^{g}}^{*}$ for the global human mesh offset. The units of orientation, translation, and vertex motion errors are degree, mm, and mm, respectively, and we quantitatively evaluate the proposed method using these three evaluation metrics. Details of the proposed metrics are provided in the supplementary material.

\begin{table}[t]
\centering
{\small
\begin{tabular}{L{1.5cm}|C{1.5cm}C{1.5cm}C{1.5cm}}
\specialrule{.1em}{.05em}{.05em}
{In / out} & {Axis-angle} & {6D} & {Quaternion} \\ 
\hline
Axis-angle & 10.48  & 10.83 & 11.15  \\ 
\cellcolor{Gray}6D & \cellcolor{Gray}9.83  & \cellcolor{Gray}10.07  & \cellcolor{Gray}10.14  \\
Quaternion & {\bf 9.46} & 9.48 & 9.91  \\
\specialrule{.1em}{.05em}{.05em}
\end{tabular}
}
\vspace*{-1mm}
\caption{\textbf{Ablation results for GMR input and output representations on Human3.6M.} The row and the column correspond to the input local pose and the output orientation motion in GMR, respectively. Numbers denote the VME.}
\label{tab:in/out representation experiment}
\vspace*{-1mm}
\end{table}

\begin{table}[t]
\centering
{\small
\begin{tabular}{L{1.8cm}|C{1.2cm}C{1.2cm}C{1.2cm}}
\specialrule{.1em}{.05em}{.05em}
{Loss type} & {OME} $\downarrow$ & {TME} $\downarrow$ & {VME} $\downarrow$ \\ 
\hline
Axis-angle & 1.05  & 6.98 & 9.51  \\ 
\cellcolor{Gray}Angular & \cellcolor{Gray}1.06  & \cellcolor{Gray}7.03  & \cellcolor{Gray}9.46  \\
Chordal & {\bf 1.01} & {\bf 6.82} & {\bf 9.28}  \\
\specialrule{.1em}{.05em}{.05em}
\end{tabular}
}
\vspace*{-1mm}
\caption{\textbf{Ablation results for orientation losses on Human3.6M.}}
\label{tab: orientation loss experiment}
\vspace*{-1mm}
\end{table}

\begin{table}[t]
\centering
{\small
\begin{tabular}{L{2.5cm}|C{1.2cm}C{1.2cm}C{1.2cm}}
\specialrule{.1em}{.05em}{.05em}
{Losses} & {OME} $\downarrow$ & {TME} $\downarrow$ & {VME} $\downarrow$ \\ 
\hline
V & 1.02 & 7.01 & 9.46  \\ 
\cellcolor{Gray}V+O & \cellcolor{Gray}{\underline{0.99}}  & \cellcolor{Gray}7.05  \cellcolor{Gray}& \cellcolor{Gray}9.34  \\
V+O+S & 1.03 & 6.88 & 9.31  \\
\cellcolor{Gray}V+O+S+T & \cellcolor{Gray}1.01 & \cellcolor{Gray}{\underline{6.82}} & \cellcolor{Gray}{\underline{9.28}} \\
\hline
V+O+S+T(4/2048) & {\bf 0.76} & {\bf 5.14} & {\bf 7.01} \\
\specialrule{.1em}{.05em}{.05em}
\end{tabular}
}
\vspace*{-1mm}
\caption{\textbf{Comparison results for adding loss components on Human3.6M.} V: Vertex loss, O: Orientation loss, S: Smoothness loss, T: Translation loss.}
\label{tab: loss component experiment}
\vspace*{-1mm}
\end{table}

\begin{table}[t]
\centering
{\small
\begin{tabular}{L{2.8cm}|C{1.2cm}C{1.2cm}C{1.2cm}}
\specialrule{.1em}{.05em}{.05em}
  & {OME} $\downarrow$ & {TME} $\downarrow$ & {VME} $\downarrow$ \\ 
\hline
w/o flip augmentation & 0.76  & 5.14 & 7.01  \\ 
\cellcolor{Gray}w/ flip augmentation & \cellcolor{Gray}{\bf 0.70}  & \cellcolor{Gray}{\bf 4.78}  & \cellcolor{Gray}{\bf 6.47}  \\
\specialrule{.1em}{.05em}{.05em}
\end{tabular}
}
\vspace*{-1mm}
\caption{\textbf{Comparison result for flip augmentation on Human3.6M.}}
\label{tab: flip augmentation experiment}
\vspace*{-1mm}
\end{table}

\subsection{Ablation Experiments}

\textbf{Analysis of GMR input and output representation.} Table~\ref{tab:in/out representation experiment} presents the quantitative comparison of nine possible combinations of 3D rotation representations for the input local pose $L$ and the output orientation motion $\Delta{A}$. The number of layers and hidden units of GRU are set to 2 and 512, respectively, in all ablation experiments for simplicity. In this experiment, the network is trained using only the vertex loss $L_{vertex}$. We conduct experiments using axis-angle, 6D~\cite{zhou2019continuity}, and unit-quaternion forms, which are widely used to represent the 3D rotation in existing human pose estimation methods. Using the 6D rotation form as the output of the network can achieve satisfactory performance due to its continuity in angular representation~\cite{zhou2019continuity}. In our GMR, however, the orientation motion has a small magnitude and causes a relatively less continuity problem than other pose estimation cases. In our experiments, the quaternion/axis-angle combination outperforms other combinations, proving that the proposed method is relatively free from discontinuity problems.

\textbf{Analysis of orientation losses.} We attempt to find the optimal orientation loss from three candidates in Sec.~\ref{subsection: loss function} to improve the GMR training. GMR is trained using the final loss function in Eq.~\ref{eq: total loss function} for fair comparison. Table~\ref{tab: orientation loss experiment} shows the quantitative comparison results. We demonstrate that chordal loss $L_{chordal}$ defined by the Frobenius norm of the $3\times3$ rotation matrix shows better performance than others. From these results, we observe that applying a loss function to the rotation matrix produces a better global motion in the proposed method. Similar to our observation, state-of-the-art human pose estimation methods~\cite{kocabas2020vibe, kolotouros2019learning} also incorporate the chordal loss. We use the chordal loss as the orientation loss according to the experimental results.

\textbf{Analysis of loss components.} The effect of each loss component is presented in Table~\ref{tab: loss component experiment}. When the orientation loss is added to the vertex loss, the orientation motion estimation performance is improved as we expected. When the smoothness loss is added, the translation and the vertex motion errors are reduced, while the orientation motion error increases. The smoothness loss forces the model to generate a smooth orientation motion, but it also causes the orientation motion to be estimated in the wrong direction. Finally, V+O+S+T outperforms V+O+S for all evaluation metrics. Although V+O+S+T shows lower performance in the orientation motion error than V+O, the effect is trivial. Therefore, we use V+O+S+T as the final loss function. Also, the performance of the GRU structure of (4, 2048) outperforms the performance of (2, 512), so we adopt the GRU structure with 4 layers and 2048 hidden units as the final model. Please note that more extensive ablation experiments for the GRU structure are provided in the supplementary material.

\textbf{Effect of flip augmentation.} The results of quantitative analysis on the effect of flip augmentation are presented in Table~\ref{tab: flip augmentation experiment}. The flip augmentation can produce physically impossible motions that can harm the performance of the proposed method. However, according to Table~\ref{tab: flip augmentation experiment}, the flip augmentation enhances the performance of all quantitative evaluation metrics. These results show that flipping many sequences in the AMASS dataset is physically plausible and thus the use of flipped sequences helps the learning of GMR by increasing the diversity of training data. Even a small number of non-reversible actions can positively affect the performance by regularizing the model.

\subsection{Comparison with Existing Method}
\label{subsection: VIBE-CAM detail}

\textbf{Baseline.} Compared with existing pose estimation methods, we present quantitative and qualitative evaluation results that show the advantages and limitations of our new framework. Specifically, we combine existing methods~\cite{kocabas2020vibe,habermann2020deepcap,raaj2019efficient}. We first reconstruct a 3D human pose and shape sequence in the human-centered coordinate system from an input video using VIBE~\cite{kocabas2020vibe}. We then obtain a 2D human pose sequence by applying the 2D human pose tracking method STAF~\cite{raaj2019efficient} to the input video. The global alignment module in DeepCap~\cite{habermann2020deepcap} computes the translation of the subject through the alignment process between 3D and 2D human poses from VIBE~\cite{kocabas2020vibe} and STAF~\cite{raaj2019efficient}, respectively. The overall procedure provides a 3D human mesh sequence in the camera coordinate system. We call this baseline VIBE-CAM and use the baseline for comparison.

\begin{table}[t]
\centering
{\small
\begin{tabular}{L{2.2cm}|C{1.4cm}C{1.4cm}C{1.4cm}} 
\specialrule{.1em}{.05em}{.05em}
Method & {OME} $\downarrow$ & {TME} $\downarrow$  & {VME} $\downarrow$   \\
\hline
VIBE-CAM & 3.88 & 49.83 & 127.07   \\ 
\cellcolor{Gray}Ours & \cellcolor{Gray}{\bf 3.67} & \cellcolor{Gray}{\bf 38.55} & \cellcolor{Gray}{\bf 120.37} \\
\hline
Ours(GT input) & 1.60 & 27.55 & 29.39 \\
\specialrule{.1em}{.05em}{.05em}
\end{tabular}
}
\vspace*{-1mm}
\caption{\textbf{Quantitative comparison between the proposed method and VIBE-CAM on the 3DPW dataset.} Ours(GT input) indicates that the ground-truth local pose is used as the input of GMR.}
\label{tab: comparion with VIBE,DeepCap,STAF on 3DPW}
\vspace*{-1mm}
\end{table}

\begin{table}[t]
\centering
{\scriptsize
\begin{tabular}{L{1.3cm}|C{0.7cm}C{0.7cm}C{0.7cm}|C{0.7cm}C{0.7cm}C{0.7cm}} 
\specialrule{.1em}{.05em}{.05em}
  & \multicolumn{3}{c}{Camera-motion-off} & \multicolumn{3}{c}{Camera-motion-on} \\
Method & {OME} & {TME}  & {VME}  & {OME} & {TME} & {VME}  \\
\hline
VIBE-CAM & {\bf 3.77} & 58.01 & 117.28 & 4.66 & 81.70 & 132.63   \\ 
\cellcolor{Gray}Ours & \cellcolor{Gray}3.80 & \cellcolor{Gray}{\bf 36.37} & \cellcolor{Gray}{\bf 105.11}   & \cellcolor{Gray}{\bf 4.01} & \cellcolor{Gray}{\bf 39.27} & \cellcolor{Gray}{\bf 108.08}   \\
\specialrule{.1em}{.05em}{.05em}
\end{tabular}
}
\vspace*{-1mm}
\caption{\textbf{Quantitative comparison between the proposed method and VIBE-CAM on the synthetic dataset.} Camera-motion-off indicates the synthetic video created without camera motion, while Camera-motion-on means the synthetic video with camera motion.}
\label{tab: comparion with VIBE,DeepCap,STAF}
\vspace*{-1mm}
\end{table}

\begin{figure}[t]
\includegraphics[width=1.0\columnwidth]{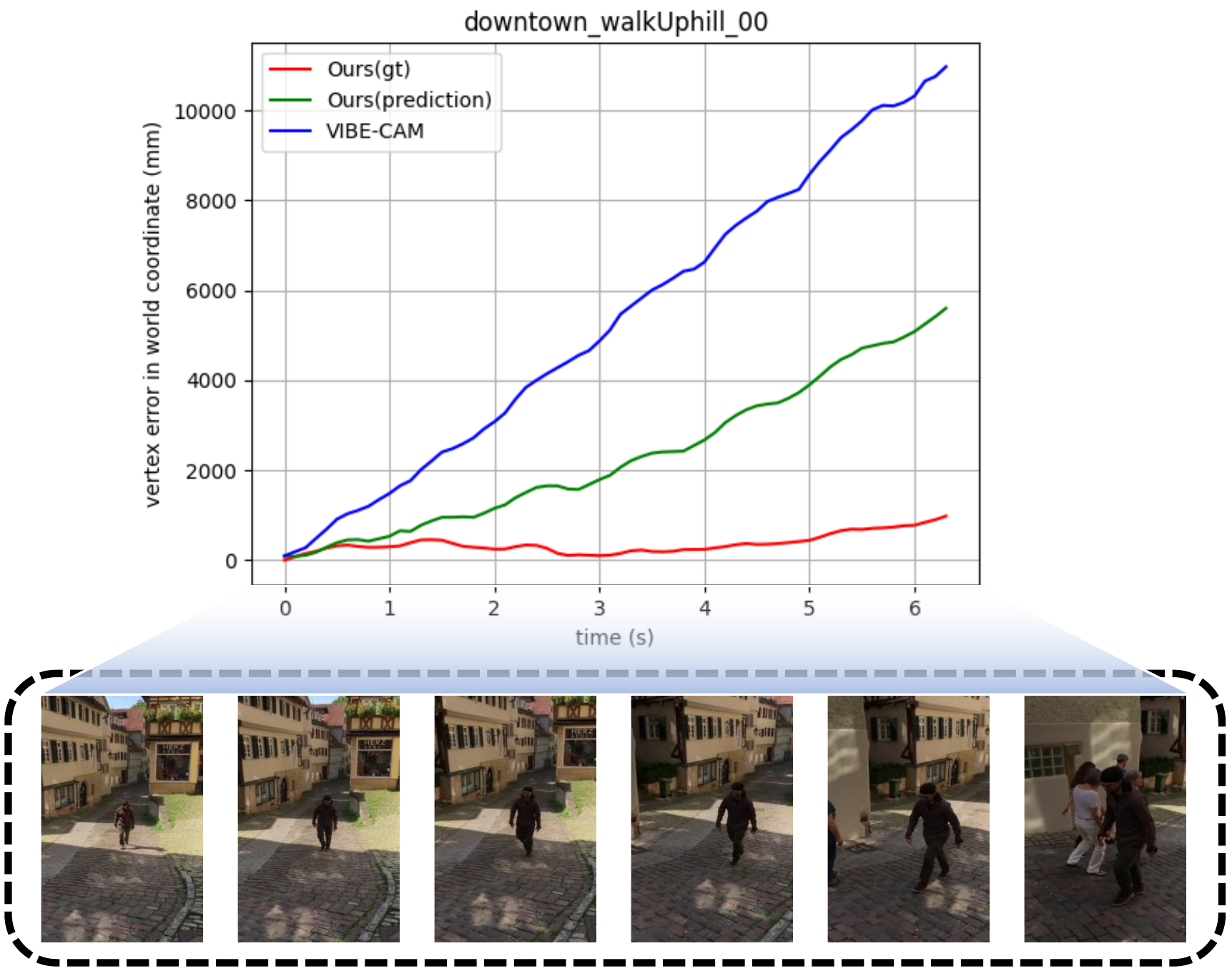}
\vspace*{-6mm}
\caption{\textbf{Vertex error over time.} The numbers in the graph represent the vertex error between the predicted human mesh and its ground-truth in the world coordinate system.}
\label{fig5}
\vspace*{-1mm}
\end{figure}


\begin{figure*}[t]
\includegraphics[width=\linewidth]{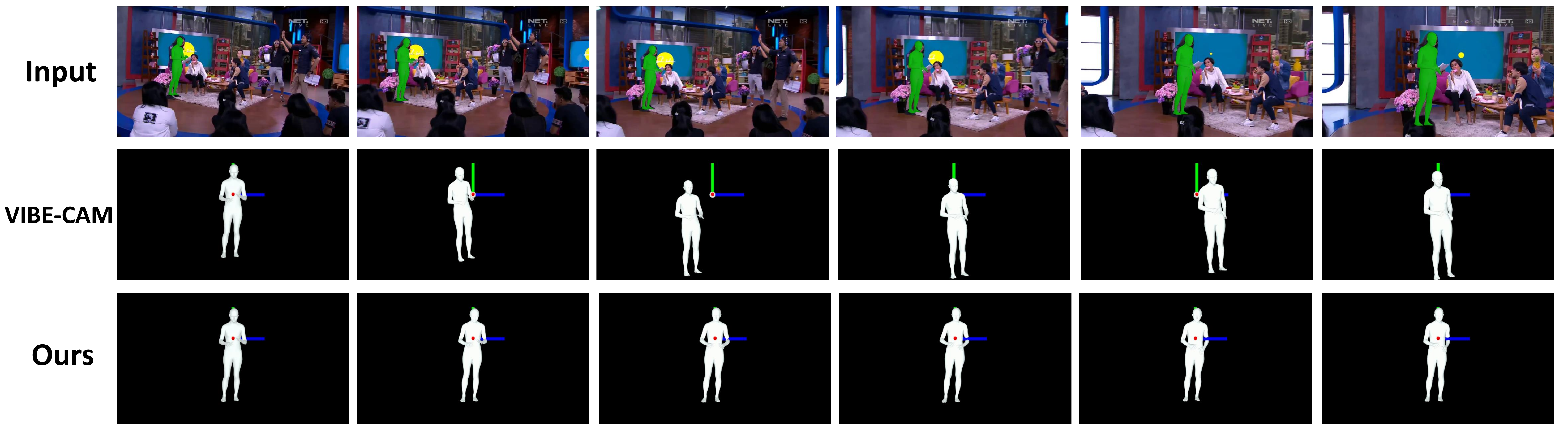}
\vspace*{-6mm}
\caption{\textbf{Qualitative comparison on the Mannequin Challenge dataset.} The proposed method provides static human poses while VIBE-CAM reconstructs unexpected global human poses with respect to the camera movement in the input video. Note that the reference coordinate systems of VIBE-CAM is aligned with that of the proposed method for easy comparison.}
\label{fig6}
\vspace*{-1mm}
\end{figure*}

\textbf{Quantitative results.} The quantitative comparison with VIBE-CAM for the 3DPW dataset is presented in Table~\ref{tab: comparion with VIBE,DeepCap,STAF on 3DPW}. The resultant global motion of VIBE-CAM is very different from the ground-truth motion because it yields global human poses in the camera coordinate system. Therefore, the proposed GMR significantly outperforms VIBE-CAM in all metrics. The results are further improved when the ground-truth local pose is used as the input of GMR. This shows that there remains a lot of room for performance improvement through better local pose estimation. The quantitative comparison results for the synthetic dataset are presented in Table~\ref{tab: comparion with VIBE,DeepCap,STAF}. The proposed framework outperforms VIBE-CAM, except for the orientation motion error, in the camera-motion-off case. However, when camera moves, VIBE-CAM reconstructs 3D human poses in the camera coordinate system, resulting in a global motion estimate much different from the ground-truth motion. Therefore, the proposed GMR significantly outperforms VIBE-CAM in all metrics. All quantitative results demonstrate that the proposed scheme provides an intrinsic global human motion independent of any camera motion embedded in the input video.

\textbf{Analysis on accumulated motion.} Fig.~\ref{fig5} shows the comparison of our reconstructed global pose with VIBE-CAM results over time. Note that the reconstructed global pose is derived from the accumulation and transformation of the global motion sequence, described in Sec.~\ref{subsection: inference}. First, the vertex error (blue line) of VIBE-CAM increases significantly because VIBE-CAM estimates the pose in the camera coordinate system. In the case of proposed method, while the vertex error still increases over time, however, the graph (green line) shows a relatively lower error than VIBE-CAM. The error increase is unavoidable because the motion error is also accumulated in the global pose reconstruction step. We believe that this error accumulation problem can be alleviated through the use of constraints, such as loop closure in methods for simultaneous localization and mapping~\cite{strasdat2011double}. The proposed method shows much lower vertex error graph (red line) when we use the ground-truth local pose in our GMR network. It demonstrates that the proposed GMR model is well-trained and effectively regresses the global motion sequence from the local pose sequence.


\textbf{Results for Mannequin Challenge dataset.} Fig.~\ref{fig6} shows an interesting result on the Mannequin Challenge dataset~\cite{li2019learning}. The dataset consists of videos that include static people in a moving camera environment, as shown in Fig.~\ref{fig6} (top row). Therefore, the 3D pose of a person reconstructed through the proposed method should not change temporally in the world coordinate system. In Fig.~\ref{fig6} (middle row), VIBE-CAM predicts the human pose in the camera coordinate system and shows unexpected human movement with respect to the camera motion in the video. In Fig.~\ref{fig6} (bottom row), however, the reconstructed human pose in our framework shows no movements as the original Mannequin Challenge scenario says. Therefore, the proposed framework effectively predicts the intrinsic human pose regardless of camera movement. Additional results on the 3DPW, synthetic, and Mannequin Challenge datasets are available in the supplementary material.

\subsection{Limitation of Proposed Method}

In this paper, we argued that the proposed method can generate global human poses defined in the world coordinate system. However, strictly speaking, it is over-claiming. For example, if a person moves inside a train running at a constant speed, then the global human pose computed by the proposed method is defined based on the running train rather than the ground on which the world coordinate system is usually based. Therefore, in the proposed method, it can only be argued that the global human pose is computed in \emph{a single coordinate system that is consistent with the overall motion of the entire sequence}. Determining whether this coordinate system is aligned to the world is impossible without calibrating the camera to the world. However, 3D human poses reconstructed in this coordinate system are still independent of camera motion and can provide valuable information for various applications. We refer to this coordinate system as the world coordinate system in this paper for convenience.

\section{Conclusion}

In this paper, a camera motion agnostic method for estimating 3D human poses in the world coordinate system is presented. Most of the 3D human pose estimation methods estimate 3D poses defined in the camera coordinate system, so it is difficult to obtain a pure human pose from a video with camera motion. To address this issue, we propose a network that generates the global motion sequence invariant to the selection of the coordinate system from the local pose sequence. Our method can reconstruct the global human mesh defined in the world coordinate system in the inference stage. We generate a pseudo ground-truth global human pose dataset from 3DPW and construct a synthetic video dataset to evaluate the proposed method. We conduct thorough experiments for quantitative and qualitative evaluation and prove the effectiveness of the proposed method.


\section*{Supplementary Material}

In this supplementary material, we provide details on the processing of the 3DPW dataset, generation of our synthetic video dataset, loss functions, and evaluation metrics and additional experimental results that could not be included in the main manuscript due to the lack of space.

\renewcommand{\thesection}{S\arabic{section}}
\setcounter{section}{0}

\begin{table*}[t]
\small
\centering
{
\begin{tabular}{L{5cm}|C{5cm}C{5cm}} 
\specialrule{.1em}{.05em}{.05em}
Sequence Name & {Frame Range}  & {Camera Motion Type}    \\
\hline
 courtyard{\_}basketball{\_}00 & 00000$.$jpg {$-$} 00467$.$jpg & Small  \\ 
 courtyard{\_}basketball{\_}01 & 00000$.$jpg {$-$} 00957$.$jpg & Small   \\
 courtyard{\_}bodyScannerMotions{\_}00 & 00000$.$jpg {$-$} 01256$.$jpg & Small  \\
 courtyard{\_}box{\_}00 & 00000$.$jpg {$-$} 01040$.$jpg & Small  \\
 courtyard{\_}captureSelfies{\_}00 & 00300$.$jpg {$-$} 00696$.$jpg & Small   \\
 courtyard{\_}golf{\_}00 & 00000$.$jpg {$-$} 00603$.$jpg & Small  \\
 courtyard{\_}rangeOfMotions{\_}00 & 00000$.$jpg {$-$} 00600$.$jpg & Small \\
 courtyard{\_}rangeOfMotions{\_}01 & 00000$.$jpg {$-$} 00586$.$jpg & Small  \\
 downtown{\_}arguing{\_}00 & 00000$.$jpg {$-$} 00897$.$jpg & Small  \\
 downtown{\_}crossStreets{\_}00 & 00000$.$jpg {$-$} 00587$.$jpg & Panning \\
 downtown{\_}runForBus{\_}00 & 00000$.$jpg {$-$} 00207$.$jpg & Linear  \\
 downtown{\_}sitOnStairs{\_}00 & 00000$.$jpg {$-$} 00477$.$jpg & Linear $\&$ Panning  \\
 downtown{\_}walkBridge{\_}01 & 00042$.$jpg {$-$} 00234$.$jpg & Panning  \\
 downtown{\_}walkDownhill{\_}00 & 00132$.$jpg {$-$} 00435$.$jpg & Panning  \\
 downtown{\_}walkUphill{\_}00 & 00000$.$jpg {$-$} 00285$.$jpg & Panning  \\
 downtown{\_}windowShopping{\_}00 & 00048$.$jpg {$-$} 00327$.$jpg & Panning   \\
 downtown{\_}windowShopping{\_}00 & 00972$.$jpg {$-$} 01542$.$jpg & Linear  \\
\specialrule{.1em}{.05em}{.05em}
\end{tabular}
}
\vspace*{-1mm}
\caption{\textbf{Details of the processed 3DPW dataset.}}
\label{tab: Details of the processed 3DPW dataset.}
\vspace*{-1mm}
\end{table*}

\begin{table*}[t]
\small
\centering
{
\begin{tabular}{L{9cm}|C{3cm}C{3cm}} 
\specialrule{.1em}{.05em}{.05em}
Sequence Name & {Number of Frames}  & {Camera Motion Type}    \\
\hline
 02{\_}03 (run/jog) & 173  & Pannning  \\ 
 08{\_}02 (walk) & 167 & Pannning   \\
 08{\_}04 (slow walk) & 484 & Pannning  \\
 08{\_}05 (walk/strid) & 298 & Pannning  \\
 13{\_}35 (climb 3 steps) & 447  & Circular  \\ 
 15{\_}02 (climb, step over, sit on, stand on stepstool) & 601 &  Circular  \\
 15{\_}10 (sit on high stool, stand up) & 598 & Circular  \\
 17{\_}08 (muscular, heavyset person's walk) &  328 & Circular  \\
 36{\_}04 (walk on uneven terrain) & 597  &  Circular \\ 
 38{\_}04 (walk around, frequent turns, cyclic walk along a line) & 596  & Circular   \\
 54{\_}01 (monkey (human subject)) & 601 &  Circular \\
 54{\_}02 (bear (human subject)) & 601 & Circular  \\
 54{\_}03 (penguin (human subject)) &  601 &  Circular \\ 
 54{\_}09 (insect/praying mantis (human subject)) & 600  &  Circular  \\
 54{\_}15 (prairie dog (human subject)) & 603 &  Circular \\
 54{\_}19 (squirrel - robotic like motion (human subject)) & 601  &  Small \\
 54{\_}26 (chicken (human subject)) & 604 &  Circular \\ 
 76{\_}03 (avoid attacker)  &  605 &  Circular  \\
 76{\_}08 (careful stepping over things) &  601  & Circular \\
 76{\_}11 (quick large steps backwards) &  513  & Circular  \\
 77{\_}14 (careful creeping) &  601  & Circular  \\ 
 77{\_}19 (limping, hurt right leg) &  600  &  Circular \\
 77{\_}31 (silent walk) & 601   &  Circular \\
 105{\_}03 (Walk8) & 513  & Circular  \\
 105{\_}05 (mummyWalk) &  601 & Circular \\ 
 105{\_}17 (QuickWalk) &  331 &  Circular  \\
 105{\_}24 (HurtLegWalk) & 601  & Circular \\
 105{\_}30 (SexyWalk) & 601  &  Circular \\
 105{\_}62 (walkFigure8) & 601  &  Circular \\ 
 111{\_}31 (Stepping up / stepping down) & 601 & Circular  \\
 111{\_}36 (Walk and carry) & 604  &  Circular \\
 120{\_}01 (Alien Turn) & 601  &  Circular \\
 120{\_}09 (Mickey sneaking walk) & 610 &  Circular \\
 127{\_}29 (Run Duck Underneath) & 209  &  Linear \\ 
 137{\_}38 (Sexy Lady Walk) & 449 &  Circular  \\
 139{\_}15 (Sneak Sideways) & 600  & Circular  \\
 139{\_}19 (Walk Wounded Leg) & 605  & Circular  \\
 139{\_}24 (Walk Wounded Leg) & 601  &  Circular \\ 
 139{\_}29 (Sneaking) & 606  &   Circular \\
 139{\_}30 (Walking) & 601  & Circular  \\
 139{\_}31 (Sneaking) & 607  & Circular  \\
 143{\_}04 (Run Figure 8) & 492 &  Circular \\
 143{\_}14 (Walk And Step Over) & 600  &  Circular \\
 143{\_}17 (Walk Up Stairs And Over) & 608 & Circular  \\ 
 143{\_}28 (Sweeping, Push Broom) & 643  &  Circular \\
 143{\_}31 (Hopscotch) & 423 & Circular  \\
 143{\_}36 (Airplane) & 600  &  Circular\\
 143{\_}39 (Walk Backwards) & 510  & Circular  \\ 
 143{\_}40 (Walk Sideways) & 601  &  Circular  \\
 143{\_}41 (Sneak) & 502 & Circular  \\
 
\specialrule{.1em}{.05em}{.05em}
\end{tabular}
}
\vspace*{-1mm}
\caption{\textbf{Details of the synthetic dataset.}}
\label{tab: Details of the synthetic dataset.}
\vspace*{-1mm}
\end{table*}

\section{Processing of 3DPW Dataset}
\label{sec:processing_of_3DPW_dataset}

The 3DPW~\cite{von2018recovering} dataset provides global human poses for evaluating the proposed method. However, the provided global poses are difficult to use for evaluation due to severe drift. For evaluation on the 3DPW dataset, we acquire camera poses from the 3DPW dataset using the existing structure-from-motion method, COLMAP~\cite{schonberger2016structure}, and use them to generate pseudo ground-truth global human poses. The 3DPW dataset provides relatively accurate 3D human poses defined in the camera coordinate system. Therefore, we convert the 3D human pose defined in the camera coordinate system into the world coordinate system using the camera pose obtained through COLMAP as follows:
\begin{equation}
\label{eq: global pose calculation using camera pose of colmap}
    \begin{bmatrix}
    R_{w} & T_{w} \\
    \mathbf{0}^{T} & 1 \\
    \end{bmatrix}
    =
    \begin{bmatrix}
    R_{col} & T_{col} \\
    \mathbf{0}^{T} & 1 \\
    \end{bmatrix}
    \begin{bmatrix}
    R_{c} & T_{c} \\
    \mathbf{0}^{T} & 1 \\
    \end{bmatrix}
    ,
\end{equation}
\begin{equation}
\label{eq: orientation calculation using camera pose of colmap}
    R_{w}=R_{col}R_{c},
\end{equation}
\begin{equation}
\label{eq: translation calculation using camera pose of colmap}
    T_{w}=R_{col}T_{c}+T_{col},
\end{equation}
where $R_{w}$ and $T_{w}$ denote the pseudo ground-truth global human pose in the world coordinate system, $R_{col}$ and $T_{col}$ denote the camera pose obtained through COLMAP, and $R_{c}$ and $T_{c}$ denote the orientation and translation of the human subject defined in the camera coordinate system. The 3DPW dataset provides camera intrinsic parameters, which can be utilized for camera calibration. Since the 3DPW dataset contains dynamic objects, simply applying COLMAP often fails to obtain successful results. Therefore, we mask out dynamic objects using the existing segmentation method, Mask R-CNN~\cite{he2017mask}, so that COLMAP extracts features only from static regions. After automatic reconstruction through COLMAP, we manually filter out sequences that fail to reconstruct successful results. Also, frames with severe drift in the reconstructed sequence are manually discarded. As a result, we obtain global human poses for 17 sequences and perform evaluations on these sequences. Table~\ref{tab: Details of the processed 3DPW dataset.} shows the details of the processed 3DPW dataset. We divide the types of camera motion into ``Small'', ``Linear'', and ``Panning''. ``Small'' indicates a sequence with little camera motion. ``Linear'' denotes the linear camera motion. And, ``Panning'' means that the camera moves horizontally around a fixed position.

\section{Creation of Synthetic Video Dataset}
\label{sec:creation_of_synthetic_video_dataset}

We construct our synthetic video dataset using general 3D animation production methods. In the Blender tool (\url{https://www.blender.org/}), we import the CMU motion BVH data. We also import a 3D human model that can generate 3D human animation sequences from the Adobe Mixamo character repository (\url{https://www.mixamo.com/}). 3D animation sequences are created by the Blender tool. Finally, we include the camera motion in animation sequences to obtain synthetic videos with the camera motion. Table~\ref{tab: Details of the synthetic dataset.} gives the details of our synthetic dataset. In addition to the camera motion present in the 3DPW dataset, we adopt the circular camera motion to construct the synthetic dataset. We observed that it is more challenging than ``Linear'' or ``Panning'' camera motions.

\begin{table*}[t]
\small
\centering
\begin{tabular}{c|ccc|ccc|ccc}
\specialrule{.1em}{.05em}{.05em}
 Layers / hidden units & \multicolumn{3}{c}{512} & \multicolumn{3}{c}{1024} & \multicolumn{3}{c}{2048} \\
 & {OME} $\downarrow$ & {TME} $\downarrow$  & {VME} $\downarrow$  & {OME} $\downarrow$ & {TME} $\downarrow$ & {VME} $\downarrow$ & {OME} $\downarrow$ & {TME} $\downarrow$ & {VME} $\downarrow$  \\
\hline
2 & 1.00 & 6.82 & 9.28 & 0.93 & 6.33 & 8.56 & 0.85 & 5.92 & 7.92   \\ 
\cellcolor{Gray}3 & \cellcolor{Gray}0.92 & \cellcolor{Gray}6.28 & \cellcolor{Gray}8.58   & \cellcolor{Gray}0.86 & \cellcolor{Gray}5.91 & \cellcolor{Gray}7.96 & \cellcolor{Gray}0.79 & \cellcolor{Gray}5.56 & \cellcolor{Gray}7.47  \\
4 & 0.85 & 5.70 & 7.74 & 0.80 & 5.48 & 7.37 & \textbf{0.76} & \textbf{5.13} & \textbf{7.00} \\
\specialrule{.1em}{.05em}{.05em}
\end{tabular}
\captionof{table}{\textbf{Ablation results for GRU structure on Human3.6M~\cite{ionescu2013human3}.}
\label{tab: gru structure}}
\end{table*}

\section{Loss Function}
\label{sec:loss_function}

Since the 3D rotation belongs to $SO(3)$, not Euclidean space, we have to carefully define the loss function to supervise the predicted orientation motion $\Delta{R_{i}}$. Hartley~\textit{et al.}~\cite{hartley2013rotation} described various distance measures that can be used for the elements of $SO(3)$. Taking them into account, we test the angular loss $L_{angular}$, the chordal loss $L_{chordal}$, and the axis-angle loss $L_{axis\text{-}angle}$, which are based on the commonly used distance measures for $SO(3)$, defined as follows:
\begin{equation}
\label{eq: angular loss}
    L_{angular}={\sum^{T}_{i=1}}{\|\log({\Delta{R}_{i}}{{\Delta{R}^{*}_{i}}}^T)\|}^{2}_{2},
\end{equation}
\begin{equation}
\label{eq: chordal loss}
    L_{chordal}={\sum^{T}_{i=1}}{\|{\Delta{R}_{i}}-{\Delta{R}^{*}_{i}}\|}^{2}_{F},
\end{equation}
\begin{equation}
\label{eq: axis-angle loss}
    L_{axis{\text -}angle}={\sum^{T}_{i=1}}{\|{\log(\Delta{R}_{i})}-{\log(\Delta{R}^{*}_{i})}\|}^{2}_{2},
\end{equation}
where ${\Delta{R}_{i}}{{\Delta{R}^{*}_{i}}}^T$, ${\Delta{R}_{i}}$, and ${\Delta{R}^{*}_{i}}$ are mapped to an axis-angle form through the logarithm map, and $*$ indicates the ground-truth. We also define the translation loss $L_{trans}$ as follows:
\begin{equation}
\label{eq: translation motion loss}
    L_{trans}={\sum^{T}_{i=1}}{\|{\Delta{T}_{i}}-{\Delta{T}^{*}_{i}}\|}^{2}_{2}.
\end{equation}
For vertex-wise loss on the reconstructed 3D mesh surface, we further define the vertex loss $L_{vertex}$ as follows:
\begin{equation}
\label{eq: vertex loss}
    L_{vertex}={\sum^{T}_{i=1}}{\sum^{N}_{j=1}}{\|{\Delta{M}^{g}_{i}[j]}-{\Delta{{M}^{g}_{i}}^{*}[j]}\|}_{1},
\end{equation}
where $\Delta{M}^{g}_{i}[j]$ denotes the $j$-th row vector of matrix $\Delta{M}^{g}_{i}$, that is, the coordinates of the $j$-th vertex, and $N=6890$ is the total number of vertices. GMR predicts the global motion, which is the temporal deviation of global poses between two adjacent frames. Therefore, instead of directly supervising the global human mesh $M^{g}_{i}$, we apply a loss function to the global human mesh offset $\Delta{M}^{g}_{i}=M^{g}([\Delta{A}_{i},L_{i}],\beta_{i},\Delta{T}_{i})$. Finally, we use the smoothness loss to generate a smooth global motion:
\begin{equation}
\label{eq: smoothness loss}
    L_{smooth}={\sum^{T-1}_{i=1}}{\|{\Delta{R}_{i}}-{\Delta{R}_{i+1}}\|}^{2}_{F},
\end{equation}
which helps to reconstruct temporally coherent global orientations in adjacent frames.


\section{Evaluation Metrics}
\label{sec:evaluation_metrics}

As far as we know, there is no metric for quantitatively evaluating the global human motion predicted by the proposed method. Therefore, we newly propose the following metrics for evaluating the proposed method. The first evaluation metric is the orientation motion error (OME) and is defined as follows:
\begin{equation}
\label{eq: orientation motion error}
    E_{orien}=\frac{1}{T}{\sum^{T}_{i=1}}{\|log({\Delta{R}_{i}^{*}}{\Delta{R}_{i}^{T}})\|}_{2},
\end{equation}
where $\Delta{R}_{i}\in{SO(3)}$ satisfies $\Delta{R}_{i}^{T}\Delta{R}_{i}=I_{3\times3}$. If the network prediction is correct, $\Delta{R}_{i}^{*}\Delta{R}_{i}^{T}=I_{3\times3}$ should hold. We transform $\Delta{R}_{i}^{*}\Delta{R}_{i}^{T}$ to $\mathbb{R}^{3}$ through the logarithm map and apply L2 norm to its result to calculate the angular error. The second evaluation metric is the translation motion error (TME) which is defined as follows:
\begin{equation}
\label{eq: translation motion error}
    E_{trans}=\frac{1}{T}{\sum^{T}_{i=1}}{\|{\Delta{T}_{i}}-{\Delta{T}_{i}^{*}}\|}_{2}.
\end{equation}
The translation motion error computes the Euclidean distance between the prediction and its ground-truth for the translation motion in $\mathbb{R}^3$. The last evaluation metric is the vertex motion error (VME) and is defined as follows:
\begin{equation}
\label{eq: vertex motion error}
    E_{vertex}=\frac{1}{TN}{\sum^{T}_{i=1}}{\sum^{N}_{j=1}}{\|{\Delta{M}^{g}_{i}[j]}-{\Delta{{M}^{g}_{i}}^{*}[j]}\|}_{2}.
\end{equation}
Since the network predicts human motion, we define the distance between the prediction and its ground-truth for the global human mesh offset as the vertex motion error. The units of orientation, translation, and vertex motion errors are degree, mm, and mm, respectively.

\section{Ablation Experiments for GRU Structure}
\label{sec:ablation_experiments}

In Table~\ref{tab: gru structure}, we present ablation results for the GRU~\cite{cho-etal-2014-learning} structure. The deeper and wider the structure of the proposed network, the better its performance without overfitting. GRU with 4 layers and 2048 hidden units shows the best performance among the candidates, so we use it as the final model.

\begin{table}[t]
\small
\centering
{\small
\begin{tabular}{L{2.2cm}|C{1.4cm}C{1.4cm}C{1.4cm}} 
\specialrule{.1em}{.05em}{.05em}
Method & {OME} $\downarrow$ & {TME} $\downarrow$  & {VME} $\downarrow$   \\
\hline
Non-sequential & 3.90 & 45.48 & 126.83   \\
\hline
Ours & {\bf3.67} & {\bf38.55} & {\bf120.37} \\
\specialrule{.1em}{.05em}{.05em}
\end{tabular}
}
\vspace*{-1mm}
\caption{\textbf{Quantitative comparison between the non sequential framework and the proposed framework on the 3DPW dataset.}}
\label{tab: comparion with non-sequential framework on 3DPW dataset}
\vspace*{-1mm}
\end{table}

\section{Reformulating VIBE}
\label{sec:reformulating_vibe}

The proposed framework can be considered as a sequential combination of the existing 3D human pose estimation network VIBE~\cite{kocabas2020vibe} and the proposed GMR. To justify our sequential framework, we perform a quantitative comparison between the non-sequential framework and the proposed framework. The non-sequential baseline can be simply constructed by reformulating VIBE to output both local pose and global motion. Unlike the proposed sequential framework, it can be learned end-to-end, which requires 2D videos and their corresponding ground-truth local poses and global motions. AMASS~\cite{mahmood2019amass} does not provide videos and the end-to-end learning is not feasible with AMASS. For the end-to-end learning, we extracted pseudo ground-truth human pose parameters from Human3.6M~\cite{ionescu2013human3} and MPI-INF-3DHP~\cite{mehta2017monocular} datasets by fitting the SMPL model to the ground-truth 3D joints in the world coordinate system using SMPLify-X~\cite{pavlakos2019expressive}. Table~\ref{tab: comparion with non-sequential framework on 3DPW dataset} shows the quantitative comparison results on the 3DPW dataset. As a result, the proposed sequential framework outperforms the non-sequential baseline for all metrics. We believe that it is because local pose estimation and global motion estimation are not highly correlated so that jointly training them makes training harder, resulting in lower performance.

\section{More Qualitative Results}
\label{sec:qualitative_results}

We provide more experimental results in the attached video, which consists of six parts. The first part shows the 3D global human pose sequences from the processed and original 3DPW datasets. As mentioned in Sec.~\ref{sec:processing_of_3DPW_dataset}, severe drift is observed for the original 3DPW dataset, but not for our processed dataset. The second part provides ablation results for flip augmentation. The third part provides qualitative comparison results with VIBE-CAM on the 3DPW dataset. In the second and third parts, VIBE~\cite{kocabas2020vibe} prediction is used as the input local pose of GMR. The fourth part provides qualitative comparison results with VIBE-CAM on the synthetic video dataset. Here, both the ground-truth local pose and the VIBE prediction are used as inputs to GMR. The results demonstrate that the proposed GMR is well-trained and effectively regresses the global human motion. The fifth part provides qualitative comparison results with VIBE-CAM on the Mannequin Challenge dataset~\cite{li2019learning}. Here, the prediction result of VIBE is used as the input to the proposed GMR. Since the images of the Mannequin Challenge dataset contain multiple subjects, we show the input video overlapped with a single target subject. The last part provides qualitative results on in-the-wild internet videos. We demonstrate the effectiveness of the proposed method through the attached video.

{\small
\bibliographystyle{ieee_fullname}
\bibliography{GMR}
}

\end{document}